\documentclass[letterpaper, 10 pt, conference]{ieeeconf}  

\IEEEoverridecommandlockouts                              

\usepackage{xcolor}
\usepackage{graphicx}
\usepackage{verbatim}
\usepackage{array}
\usepackage{tabularx}
\usepackage{makecell}
\usepackage{caption}
\usepackage{subcaption}
\usepackage{amsmath}
\usepackage{booktabs}
\usepackage{color}
\usepackage{mathtools}

\usepackage{mwe}
\usepackage{tabularx}
\usepackage{tikz}
\usepackage{pgfplots}
\usetikzlibrary{patterns}
\usepackage{tabularray}
\UseTblrLibrary{booktabs}

\usetikzlibrary{shapes,decorations,arrows,calc,arrows.meta,fit,positioning}
\tikzset{
    -Latex,auto,node distance =1 cm and 1 cm,semithick,
    state/.style ={ellipse, draw, minimum width = 0.7 cm},
    point/.style = {circle, draw, inner sep=0.04cm,fill,node contents={}},
    bidirected/.style={Latex-Latex,dashed},
    el/.style = {inner sep=2pt, align=left, sloped}
}
\pgfplotsset{width=6cm, compat=1.6}

\title{\LARGE \bf
Optimizing Warfarin Dosing Using Contextual Bandit: An Offline Policy Learning and Evaluation Method
}

\author{Yong Huang$^{1, *}$, Charles A. Downs$^{3}$, and Amir M. Rahmani$^{1,2,4}$ 
\thanks{$^{1}$Dept. of Computer Science, University of California, Irvine. 
        $^{2}$School of Nursing, University of California, Irvine. 
        $^{3}$School of Nursing and Health Sciences, University of Miami. 
        $^{4}$Institute for Future Health (IFH), University of California, Irvine. 
        {\tt\small (*correspondence email: yongh7@uci.edu)}}%
}

\begin{document}
\bstctlcite{IEEEexample:BSTcontrol}

\maketitle
\thispagestyle{empty}
\pagestyle{empty}

\begin{abstract}
Warfarin, an anticoagulant medication, is formulated to prevent and address conditions associated with abnormal blood clotting, making it one of the most prescribed drugs globally. However, determining the suitable dosage remains challenging due to individual response variations, and prescribing an incorrect dosage may lead to severe consequences. Contextual bandit and reinforcement learning have shown promise in addressing this issue. Given the wide availability of observational data and safety concerns of decision-making in healthcare, we focused on using exclusively observational data from historical policies as demonstrations to derive new policies; we utilized offline policy learning and evaluation in a contextual bandit setting to establish the optimal personalized dosage strategy. Our learned policies surpassed these baseline approaches without genotype inputs, even when given a suboptimal demonstration,  showcasing promising application potential.
\end{abstract}

\section{INTRODUCTION}

Warfarin, introduced in the 1950s, is designed to prevent and manage conditions related to abnormal blood clotting, such as deep vein thrombosis, pulmonary embolism, and stroke \cite{rettie2006pharmocogenomics}. It is estimated to be prescribed over 11 million times in the United States in 2021 \cite{Sean}. Despite its extensive use, determining the appropriate dosage poses a challenge due to variations in individual responses. Factors like age, weight, genetics, and liver and kidney function, among others, play pivotal roles in the drug's metabolism, making identification of the optimal dosage challenging \cite{kuruvilla2001review}.

Choosing an incorrect warfarin dosage can have dangerous consequences, leading to life-threatening excessive bleeding. Consequently, there is a growing interest in developing policies for determining the optimal dose. In practice, clinicians typically initiate treatment with an initial dosage, then closely monitor the patient's responses and make dosing adjustments. This iterative process continues until the optimal therapeutic dosage is identified, typically taking weeks \cite{garcia2011warfarin}.

Ideally, identifying the correct dose without requiring an extensive exploration procedure would save critical time. Numerous efforts have been proposed to predict the optimal initial dosage. For instance, the Warfarin Clinical Dosing Algorithm (WCDA) \cite{international2009estimation}, a linear model, considers variables like weight, age, and medications. Similarly, the Warfarin Pharmacogenetic Dosing Algorithm (WPDA) is another linear model that introduces additional genotype inputs \cite{international2009estimation}.

These simplistic linear models, however, often fall short of capturing the intricate relationships between variables, resulting in only marginally satisfactory success rates for predicting the correct therapeutic dose. Furthermore, the WPDA algorithms necessitate genotype inputs for more accurate predictions, posing challenges in their application to a broader population where genotype information may be unavailable. This underscores the need to develop more advanced approaches to assist clinicians in decision-making when determining the appropriate dosage.

Recent advancements in contextual bandit and reinforcement learning have demonstrated promising results in enhancing decision-making across various applications \cite{huang2022reinforcement, raghu2017deep}, and several pilot studies have attempted to use online approaches for optimizing warfarin dosing, achieving promising results \cite{vats2021estimation, xiao2019online}. However, a critical limiting factor of online approaches is their dependence on generating new data through active exploration. In other words, online learning approaches typically require generating data from the policy being improved in the learning process, rendering them less applicable in domains like healthcare, where safety and ethics are critical, and building simulations can be challenging.

Alternatively, there is a growing interest in healthcare for offline reinforcement learning and bandits \cite{gottesman2018evaluating}. Unlike online approaches, offline methods leverage historical observational data only, typically collected by another policy, and attempt to derive a new policy from it. Offline approaches lend themselves to healthcare applications where historical observational data is widely available. The offline learning nature frees us from building a simulation environment for exploration, which is challenging and complicated to extend to more complex decision-making problems, such as sepsis management. Furthermore, recent theoretical work has demonstrated that learning a policy merely from observational data that outperforms the demonstration is possible \cite{rashidinejad2021bridging}.

In this study, we frame the Warfarin dosing problem within offline contextual bandit, aiming to maximize expected rewards by identifying a high-quality dosage policy. We utilize historical observational data from baseline policies, employing offline learning algorithms to derive improved policies. Given the limited research on Off-policy evaluation (OPE) methods, we assess results using three representative OPE estimators. Our learned policy outperforms demonstration policies, even when the baseline policy is suboptimal (e.g., random dosage policy). This highlights the effectiveness of our approach to Warfarin dosing.


Our contributions are three folds:

\begin{itemize}
    \item To the best of our knowledge, this work is the first to propose a machine-learning dosing model in an offline fashion.
    \item Our novel policy is capable of either matching or surpassing the effectiveness of existing clinical baselines, and it achieves this without requiring genotype data, thus enhancing its scalability for practical implementation.
    \item We have explored the validity and reliability of the OPE methods within clinical settings, offering practical empirical evidence to guide the application and selection of these estimators.
\end{itemize}

\section{Methods}
\subsection{Dataset}
We utilize a publicly accessible patient dataset curated by the Pharmacogenomics Knowledge Base (PharmGKB) \cite{barbarino2018pharmgkb}. This dataset comprises information from 5700 patients who underwent warfarin treatment across the globe. Each patient's profile in this dataset includes various features such as demographics (e.g., gender, race), background details (e.g., height, weight, medical history), and phenotypic and genotypic information. Additionally, the dataset includes the actual patient-specific optimal warfarin doses determined through the physician-guided dose adjustment process over several weeks. We categorized the therapeutic doses into low (less than 21 mg/week), medium (21-49 mg/week), and high (more than 49 mg/week) following conventions \cite{international2009estimation}.

\subsection{Problem Formulation}
\begin{figure}[t]
\includegraphics[width=0.49\textwidth]{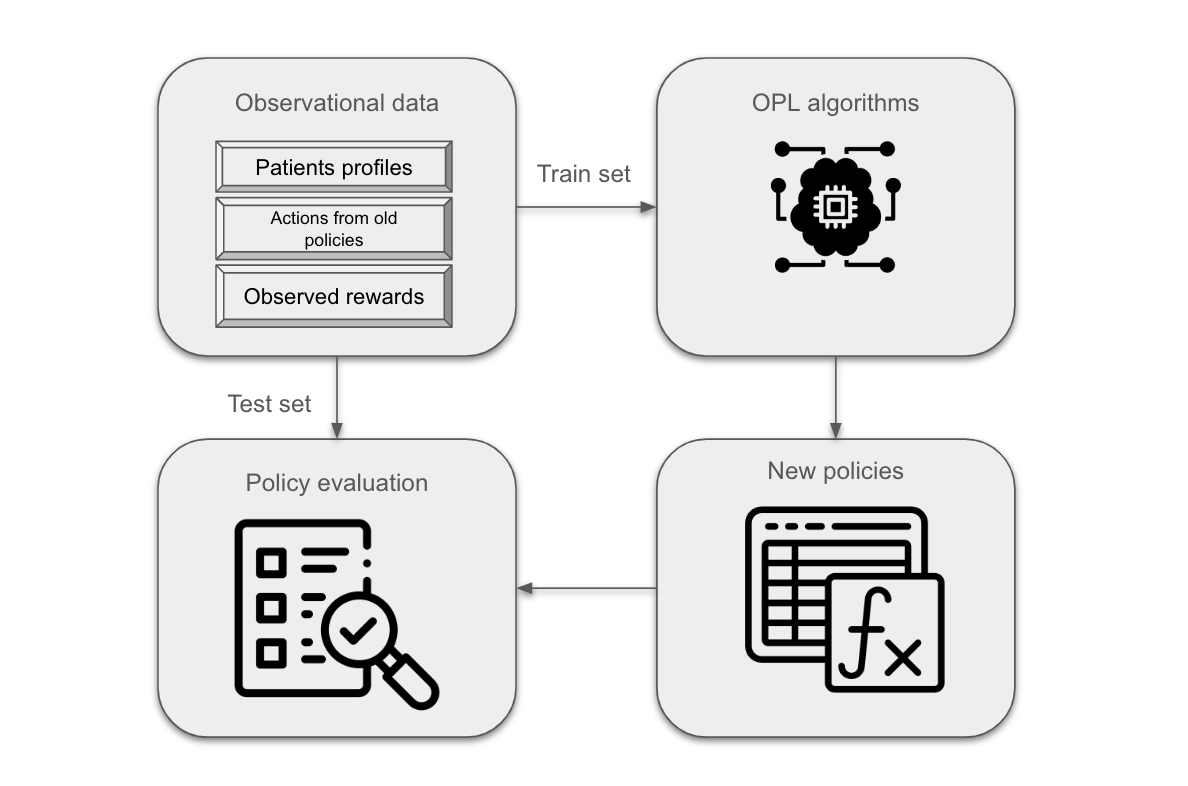}

\caption{Workflow of offline learning and evaluation, an essential distinct between contextual bandit and supervised learning is that the ground truth optimal action is not revealed to learning and evaluation algorithms, making it close to real-world decision-making problems where the outcome associated with the optimal action may be counterfactual and unavailable in observational data.}
\centering
\label{fig:diag}
\end{figure}

We formulate the problem of optimal warfarin dosing under the contextual bandit framework. The Dataset $\mathcal{D}$ consists of $\mathcal{N}$ patients, and for each patient, we observe its feature vector $\mathcal{X} \in \mathcal{R}^d$. This represents the available knowledge about the patient, also known as the context in bandit literature, and it is used to help determine actions $\mathcal{A}$ chosen by the policy $f$, which has access to $\mathcal{K}$ actions/arms where the action represents the warfarin dosage to provide to the patient. As defined in the previous section, we have $\mathcal{K} = 3$ arms corresponding to low, medium, and high warfarin doses.
If the algorithm identifies the correct dosage for the patient, the reward $\mathcal{R}$ is 1. Otherwise, a reward of 0 is received. The objective is to optimize the policy such that the action selected by the policy has a maximum expected reward $E[\mathcal{R}]$. To demonstrate better how offline contextual bandit works, we present the high-level idea of the workflow in Figure \ref{fig:diag}.

\subsection{Offline Policy Learning}
We implemented two offline policy learning (OPL) algorithms, the Offset Tree and the doubly robust estimator \cite{beygelzimer2009offset, dudik2011doubly}, to learn the optimal policy. The offset tree algorithm simplifies the original problem, a censored label classification problem, to binary importance-weighted classification, which can be implemented using any binary classifier, such as logistic regression. Conversely, the doubly robust (DR) estimators require two estimates: one for the reward, providing an estimate of the reward each arm will give, and another for the probability, offering an estimate of the probability that the policy collecting the data assigned to each arm chosen for each observation, which is used to weight the observed rewards. As long as one of the estimates is well-specified, the DR estimator is unbiased and, hence, doubly robust.

\subsection{Offline Policy Evaluation}
Although the true counterfactual outcome of every possible action is known under our specific reward formulation, making it possible to evaluate the performance of the proposed policy directly, it is worthwhile to investigate it through the lens of OPE. In a more general scenario, the counterfactual outcome is typically unobservable, and OPE becomes the only tool to quantify policy quality. Given that limited work has been done to assess these OPE methods empirically \cite{voloshin2019empirical}, the Warfarin problem provides an excellent opportunity to help us understand how well OPE works by comparing the estimates with the oracle ground truth. Therefore, we perform OPE to estimate the expected reward of the proposed policy and compare these estimates with the ground truth expected reward.

At a high level, Off-policy evaluation (OPE) offers a statistical framework for estimating the value function of a specific target policy, typically utilizing information from another policy that generated the data. We implemented three policy evaluation methods: 

\begin{itemize}
    \item \textbf{Rejection Sampling \cite{li2010contextual}}: Reject sampling is an unbiased estimator that estimates the performance of a policy by collecting some data on which actions are chosen at random, a reject sampling scheme is performed later on the policy to be evaluated, namely, if the action agrees with the observation from random action, it is kept and it is rejected otherwise.
    \item \textbf{Doubly Robust \cite{dudik2011doubly}}: Doubly Robust in evaluation works the same way as described in the learning section.
    \item \textbf{Normalized Capped Importance Sampling \cite{gilotte2018offline}} Normalized Capped Importance Sampling estimates rewards of arm choices of a policy from data collected from the demonstration policy, making corrections by weighting rewards according to the difference between the estimations of the evaluation policy and demonstration policy over the actions that were chosen.
\end{itemize}

\section{EXPERIMENTS AND Discussion}
We pre-processed the dataset, which involved normalizing continuous variables and imputing missing values as necessary. For instance, missing weights were imputed by using the average value across the dataset. The dataset was then divided into an 80 percent training set and a 20 percent test set.

We implemented three baseline policies. The first policy is a random dosage policy where every patient is assigned a random dosage of warfarin. The other two policies are WCDA and WPDA algorithms. The rewards associated with the actions selected by these policies were computed. For each baseline policy, the data tuples consisting of $\langle X, A, R \rangle$  were used in the training of offline learning algorithms to derive new policies. Subsequently, we calculated the expected rewards of these learned new policies on the test set and compared them with the expected rewards of baseline policies that were used as demonstration, the OPL algorithm and OPE methods are implemented using contextual bandit library in python \cite{cortes2018adapting}.

\begin{figure}

\begin{subfigure}[a]{1.0 \textwidth}
\includegraphics[width=0.5\textwidth]{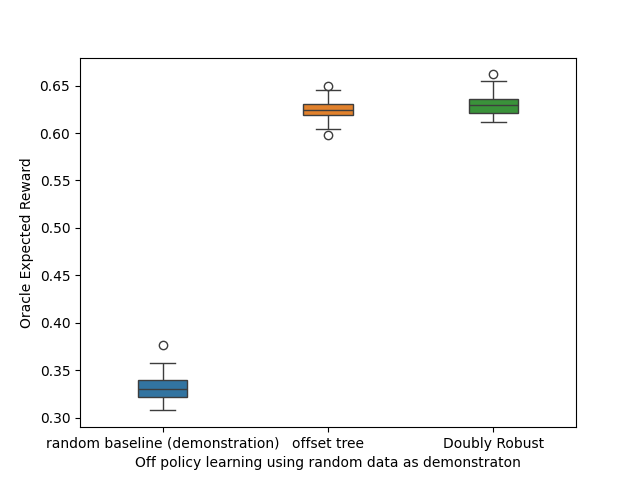}
\end{subfigure}\\[-1ex]

\begin{subfigure}{\textwidth}
\includegraphics[width=0.5\textwidth]{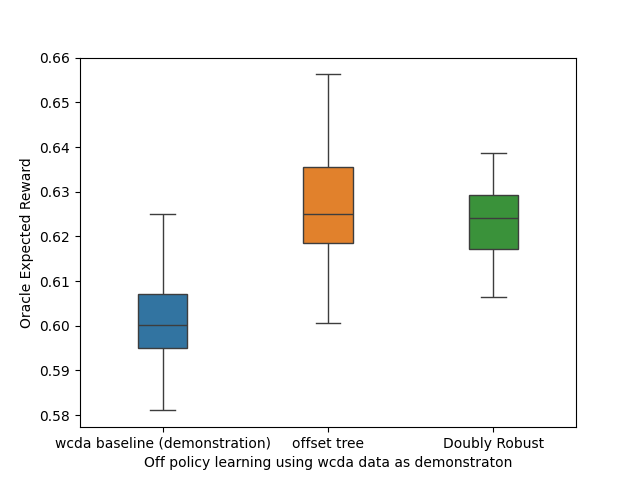}
\end{subfigure}\\[-1ex]

\begin{subfigure}{\textwidth}
\includegraphics[width=0.5\textwidth]{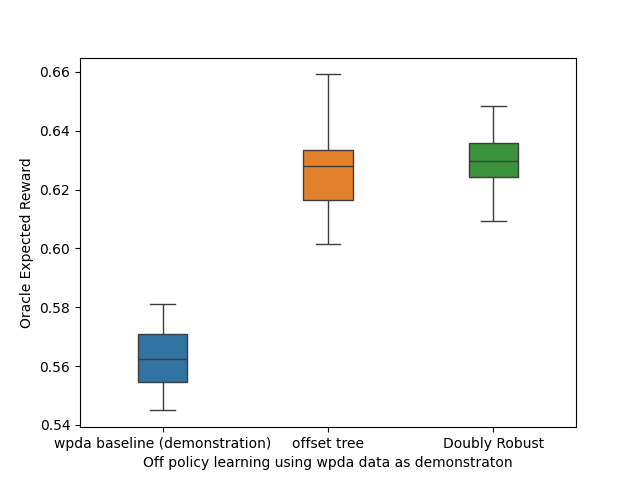}
\end{subfigure}

\caption{The expected reward of thirty experiments on test sets is presented in a boxplot. In each subfigure, offset tree and doubly robust estimator learn from a corresponding old policy.}
\label{fig2}

\end{figure}

To account for the randomness introduced by the train-test split, we performed the splitting process 30 times with different random seeds and reported the results of these 30 experiments. The offline learning results are presented in Figure \ref{fig2}; we compared the oracle reward on the left-out test set of two OPL methods against the baseline demonstration which they are learned from. Our newly learned policies outperformed their respective demonstrations by a large margin, notably when presented with the random policy data, a suboptimal policy, OPL algorithms still managed to learned a policy that matches other baselines. Furthermore, both offset tree and doubly robust performance are similar regardless what observational data they learned from, we speculate that this may be the most ideal results we could achieve given the available features collected. Importantly, the OPL algorithms showcased superior performance without requiring genotype inputs, making them more practical for deployment.

\begin{table}[!htbp]
\centering
\caption{OPE estimation of rewards and oracle rewards of new policies learned from random policy demonstration}
\begin{booktabs}{
  colspec = {crr},
  cell{1}{2,4} = {c=2}{c}, 
}
\toprule
 \textbf{OPE}     &   OT learned policy &      &  DR learned policy &      \\
\midrule
    Reject Sampling & 0.622 (0.545, 0.660) &     &0.625 (0.578, 0.660)  &  \\
    DR  & 0.619 (0.552, 0.661)   &  & 0.625 (0.577, 0.660)  &  \\
    NCIS  &  0.331 (0.308, 0.376)  &   &  0.331 (0.308, 0.376)&  \\
\midrule 
    oracle &  0.624 (0.598, 0.650)& & 0.631 (0.612, 0.662)& \\
\bottomrule
\end{booktabs}
\label{tab1}
\end{table}

\begin{table}[!htbp]
\centering
\caption{OPE estimation of rewards and oracle rewards of new policy learned from WCDA policy demonstration}
\begin{booktabs}{
  colspec = {crr},
  cell{1}{2,4} = {c=2}{c}, 
}
\toprule
 \textbf{OPE}     &   OT learned policy &      &  DR learned policy &      \\
\midrule
    Reject Sampling  & 0.650 (0.623, 0.676)  &  &  0.639 (0.618, 0.656) &  \\
    DR  &  0.651 (0.600, 0.890) &  &  0.633 (0.610, 0.653)&  \\
    NCIS  &  0.601 (0.581, 0.625)  &  & 0.601 (0.581, 0.625)  &  \\
\midrule 
    oracle &  0.627 (0.601, 0.656) & & 0.623 (0.606 ,0.639)& \\
\bottomrule
\end{booktabs}
\label{tab2}
\end{table}

\begin{table}[!htbp]
\centering
\caption{OPE estimation of rewards and oracle rewards of new policy learned from WPDA policy demonstration}
\begin{booktabs}{
  colspec = {crr},
  cell{1}{2,4} = {c=2}{c}, 
}
\toprule
 \textbf{OPE}     &   OT learned policy &      &  DR learned policy &      \\
\midrule
    Reject Sampling  &  0.652 (0.633, 0.677) &  & 0.639 (0.620, 0.663) &  \\
    DR  &  0.737 (0.444, 1.000) &  & 0.631 (0.611, 0.656)  &  \\
    NCIS  &   0.562 (0.545, 0.581) &   & 0.562 (0.545, 0.581) &  \\
\midrule 
    oracle & 0.626 (0.602, 0.659) & &0.629 (0.609, 0.648) & \\
\bottomrule
\end{booktabs}
\label{tab3}
\end{table}

Another task of interest in this work is to examine the empirical performance of OPE methods on newly proposed policies. We implemented a logistic regression to acquire the distribution over actions generated by the new policy, which is used to compute reward estimates in doubly robust OPE and NCIS OPE, and Table \ref{tab1}, \ref{tab2},, \ref{tab3} detail our experimental results on OPEs. We presented mean rewards estimation from the 30 experiments with lower and upper bounds and the actual oracle rewards. It is worth noting that the NCIS appears to deviate the most from the ground truth, while reject sampling is empirically the best choice. A potential reason for more significant errors in NCIS estimation may stem from the inaccurate importance weights we calculated using logistic regression; reject sampling typically works well when the sample size is large enough. Additionally, doubly robust methods work best with continuous actions, and the same dependency on importance weight computation makes it less practical for real-world applications. 

In conclusion, we demonstrated the feasibility of deriving an improved policy that outperforms its demonstrations using historical observational data. We empirically evaluated three representative OPE methods in a bandit setting. A primary limitation of this study is that the feature space of collected observational data may not be sufficiently representative, potentially hindering the OPL from achieving optimal performance due to unobserved confounding factors. Additionally, the empirical OPE evaluation results may not generalize to other use cases, especially considering the variability in the properties of these estimators in reinforcement learning settings. We intend to address these issues in future research.

\addtolength{\textheight}{-12cm}   





\bibliographystyle{IEEEtran}
\bibliography{IEEEabrv,ref}

\end{document}